\documentclass[letterpaper, 10 pt, journal, twoside]{ieeetran}
\usepackage{amsmath,amssymb,amsfonts}
\usepackage[pdftex]{graphicx}
\usepackage{epsfig}
\usepackage{mathptmx}
\usepackage{times}
\usepackage{amsmath} 
\usepackage{amssymb} 
\usepackage{mathtools}
\usepackage{dblfloatfix}
\usepackage{color}
\usepackage{svg}
\usepackage{booktabs,tabularx}
\usepackage{optidef}
\usepackage[font=small,skip=3pt]{caption}
\usepackage{float}
\usepackage[normalem]{ulem}
\usepackage{xcolor}
\usepackage{multirow}
\usepackage{comment}
\usepackage{float}
\usepackage{balance}
\usepackage{cite}
\usepackage{bm}
\usepackage{titlesec}
\usepackage{hyperref}
\usepackage{adjustbox}
\usepackage{booktabs}

\newcommand{\dma}[1]{{\color{black}#1}}
\newcommand{\dl }[1]{{\color{black}#1}}

\title{Compliant Fins for Locomotion in Granular Media}

\author{Dongting Li$^{1}$, Sichuan Huang$^{2}$, Yong Tang$^{2}$, Hamidreza Marvi$^{3}$, Junliang Tao$^{2}$ and Daniel M. Aukes$^{1}$

\thanks{Manuscript received: January, 8, 2021; Revised March, 6, 2021; Accepted May, 4, 2021.}

\thanks{This paper was recommended for publication by Editor Kyu-Jin Cho upon evaluation of the Associate Editor and Reviewers' comments.}

\thanks{This material is based on the work primarily supported by the National Science Foundation (NSF) under NSF Award Number CMMI-1841574.  Any opinions, findings and conclusions, or recommendations expressed in this material are those of the author(s), and do not necessarily reflect those of the NSF. \textit{(Corresponding author: Daniel M. Aukes)}}

\thanks{$^{1}$D. Li and D. M. Aukes are with the Polytechnic School, Ira A. Fulton Schools of Engineering, Arizona State University, Mesa, AZ 85212, USA, {\tt \footnotesize{dongting@asu.edu; danaukes@asu.edu}}.

$^{2}$S. Huang, Y. Tang and J. Tao are with School of Sustainability and the Built Environment, Center for Bio-mediated and Bio-inspired Geotechnics, Arizona State University, Tempe, AZ, 85287, USA, {\tt \footnotesize{shuang64@asu.edu; ytang116@asu.edu; jtao25@asu.edu}}.

$^{3}$H. Marvi is with School for Engineering of Matter, Transport and Energy, Ira A. Fulton Schools of Engineering, Arizona State University, Tempe, AZ, 85287, USA,
{\tt \footnotesize{hmarvi@asu.edu}}
}
\thanks{Digital Object Identifier (DOI): see top of this page.}
}

\begin{document}
\markboth{IEEE Robotics and Automation Letters. Preprint Version. Accepted May, 2021}
{Li \MakeLowercase{\textit{et al.}}: Compliant Fins for Locomotion in GM}

\maketitle

\begin{abstract}
In this paper, we present an approach to study the behavior of compliant plates in granular media and optimize the performance of a robot that utilizes this technique for mobility. From previous work and fundamental tests on thin plate force generation inside granular media, we introduce an origami-inspired mechanism with non-linear compliance in the joints that can be used in granular propulsion. This concept utilizes one-sided joint limits to create an asymmetric gait cycle that avoids more complicated alternatives often found in other swimming/digging robots. To analyze its locomotion as well as its shape and propulsive force, we utilize granular Resistive Force Theory (RFT) as a starting point. Adding compliance to this theory enables us to predict the time-based evolution of compliant plates when they are dragged and rotated. It also permits more rational design of swimming robots where fin design variables may be optimized against the characteristics of the granular medium. This is done using a Python-based dynamic simulation library to model the deformation of the plates and optimize aspects of the robot's gait. Finally, we prototype and test robot with a gait optimized using the modelling techniques mentioned above.
\end{abstract}
\begin{IEEEkeywords}
Soft Robot Materials and Design;
Modeling, Control, and Learning for Soft Robots;
Biologically-Inspired Robots
\end{IEEEkeywords}

\section{introduction}
\IEEEPARstart{T}{his} paper studies the role compliance can play in digging applications; more specifically, we study how flexible flat plates may be programmed with nonlinear compliance in order to preferentially locomote through granular media. This is done by utilizing 3D-printed flexible materials in the fin design, and is inspired by animal locomotion.

Animals' ability to locomote through a variety of media like sand, dirt, and fluid is a complex product of the force interactions between the musculo-skeletal systems of these animals and the medium that surrounds them \cite{hosoi_beneath_2015, lauder_fish_2007, dickinson_how_2000, lauga_hydrodynamics_2009}.
A well known concept from literature, known as the ``Scallop Theorem", states that if a swimmer in a low Reynolds number fluid environment performs a reciprocal behavior, zero net movements will be generated \cite{purcell_life_1977}.
Indeed, for a number of more complex organisms such as bony fish, the motion of fins, spines, and other active subsystems have been observed to proceed through non-reciprocating trajectories such as rowing, cupping, and undulation\cite{lucas_bending_2014, dickinson_how_2000}.
For simpler organisms, however, including the body's compliance such as the ``flexible oar'' may serve as a straightforward approach to breaking symmetry \cite{purcell_life_1977}.

Driving flexible filaments or foils to swim, propel or generate force in a viscous fluidic environment has been demonstrated as an efficient way to generate non-reciprocating motion \cite{purcell_life_1977}. The motion of ``elastic swimmers" with rigid bodies and slender, elastic tails have been described, using the compliance of the tail section to solve for the swimming kinematics such as shape, velocities and force balancing in the case of small amplitude oscillations \cite{lauga_floppy_2007}.
In another example, a flexible flapping model using a torsional spring as a compliant element has been used to study the locomotion of slender bodies in viscous fluids and granular media \cite{peng_propulsion_2017}; in this study, however, the compliance in the proposed wing-flapping mechanism exists only at the proximal joint of a rigid beam, rather than distributed along the beam as in an Euler-Bernoulli formation. In a second paper by the same group, the stiffness of a filament was varied along its length, providing new design principles for maximizing the propulsion of micro-swimmers\cite{peng_maximizing_2017}. Finally, the study of flexible plates and foils has been extended to frictional environments such as granular media with numerical solutions for the static force balance and curvature of continuum plates moving through soil\cite{wang_dynamics_2018}. These papers serve as a theoretical starting point for our current work.

Researchers have also developed a variety of swimming or digging robots inspired by this biological phenomenon to create non reciprocal motion trajectories. This includes high degree-of-freedom mechanisms\cite{mazouchova_flipper-driven_2013, marvi_sidewinding_2014} or pneumatic chambers\cite{ortiz_soft_2019,https://doi.org/10.1002/aisy.201900183,Tao_2020,doi:10.1061/(ASCE)GT.1943-5606.0002177}. Without the musculo-skeletal systems found in many digging/swimming animals, there are few simple techniques for thoughtfully adding compliance to mimic the natural swimming and digging capabilities found in nature. One natural place to add compliance and break the symmetry of a simple flapping motion is in leg or fin-like digging appendages.

Flexible and compliant materials have been used to mimic the biological gaits of digging animals.
\dl{Shoele and Zhu showed nonuniform flexibility in insect wing designs
could lead to higher lift force in air with low energy consumption comparing to rigid wings \cite{doi:10.1063/1.4802193}. Nian et al \cite{NIAN2020105944} analyzed a flexible wing with a one-sided stop at the joint. In aerodynamics, using asymmetry in the deformation between upward and downward \dma{has been shown to} increase the wing performance with higher force and lower energy consumption. \cite{tummala_design_2013,mueller_incorporation_2009}.
In granular media, using asymmetric fin designs for propulsion has been less explored.}
Russell developed a burrowing robot \cite{russell_crabot_2011} inspired by the Mole Crab, an animal that maximizes thrust in its power stroke and minimizes drag in its recovery stroke through careful configuration of its leg pairs \cite{trueman_mechanism_1970}. This robot utilizes flexible sheets and a rigid stop in the fin design to break symmetry and permit forward motion in sand. Inspired by this concept, we seek in this paper to find similar approaches that utilize nonlinear compliance to further simplify this basic design. We seek to use granular resistive force theory, coupled with the knowledge of system compliance, as a way to better understand and improve such designs.

We thus propose an origami inspired design \cite{onal_towards_2011, 6266749, wood_first_2008,tolley_self-folding_2014} in which rigid plates are connected by compliant joints to permit bending in one direction. These joints are fabricated out of two layered materials and can effectively act as a one-sided joint limit, as in \cite{doi:10.1089/soro.2019.0156}. The difference in stiffness between forward and reverse motion is then leveraged to break the symmetry of reciprocating motor inputs.

To model our digging system, we have selected granular Resistive Force Theory (RFT) \cite{maladen_undulatory_2009}, which has been used to model the propulsive forces of organisms with low Reynolds numbers \cite{gray_propulsion_1955} as well as used in the design, analysis, and optimization of robots \cite{maladen_mechanical_2011, goldman_colloquium_2014, li_terradynamics_2013, zhang_effectiveness_2014}.
Granular RFT has been integrated into a dynamics simulation package called \dl{Pynamics}\footnote{\url{https://github.com/dli-sys/Pynamics_demo\_for\_compliant_fin\_ral}}, which is a Python based dynamics simulation library that uses Kane's method to derive symbolic equations of motion \cite{sharifzadeh_integrated_2018,9369911}. We implement the RFT-based force, material properties, experimental parameters into \dl{Pynamics} to replicate the system and optimize the design. Other alternative, such as Discrete Element Method (DEM)
\cite{rapaport_art_2004} coupled with Finite Element Method (FEM)-based simulations \cite{michael_demfem_2015, tu_sequential_2017, han_combined_2000} are potentially suitable to predict the force interactions between particles and compliant intruders but incur high computational costs and take much longer, reducing their usefulness in optimization-based approaches.

The rest of this paper is organized as follows: 
in Section \ref{carft}, we derive static force balance equations based on RFT and apply it rigid elements connected by compliant joints. This compliance-augmented theory is then applied later to study fin design and the concept of ``Effective Flapping'' discussed in later sections. 
In Section \ref{section2b}, we introduce a new kind of fin inspired by origami and compare it to a fully compliant plate.
We subsequently describe a series of dragging and rotating tests performed to validate our model in Section \ref{Modelverify} and demonstrate how symmetry can be broken with our design in Section \ref{actuator_experiments} and \ref{effective_flapping}.
We then use \dl{Pynamics} to model the fin dragged through soil and match it with the result of our experimental dragging tests in Section \ref{Pynamics}.
Finally, a robotic prototype based on our concept is presented in Section \ref{Robot_tests}. Using the parameters from our plate dragging simulation, we model the robot in \dl{Pynamics} and optimize its gait for better efficiency.

\section{Test Setup and Theory}
\subsection{Test Setup}
\begin{figure}[h]
\centering
\includegraphics[width =0.9\columnwidth]{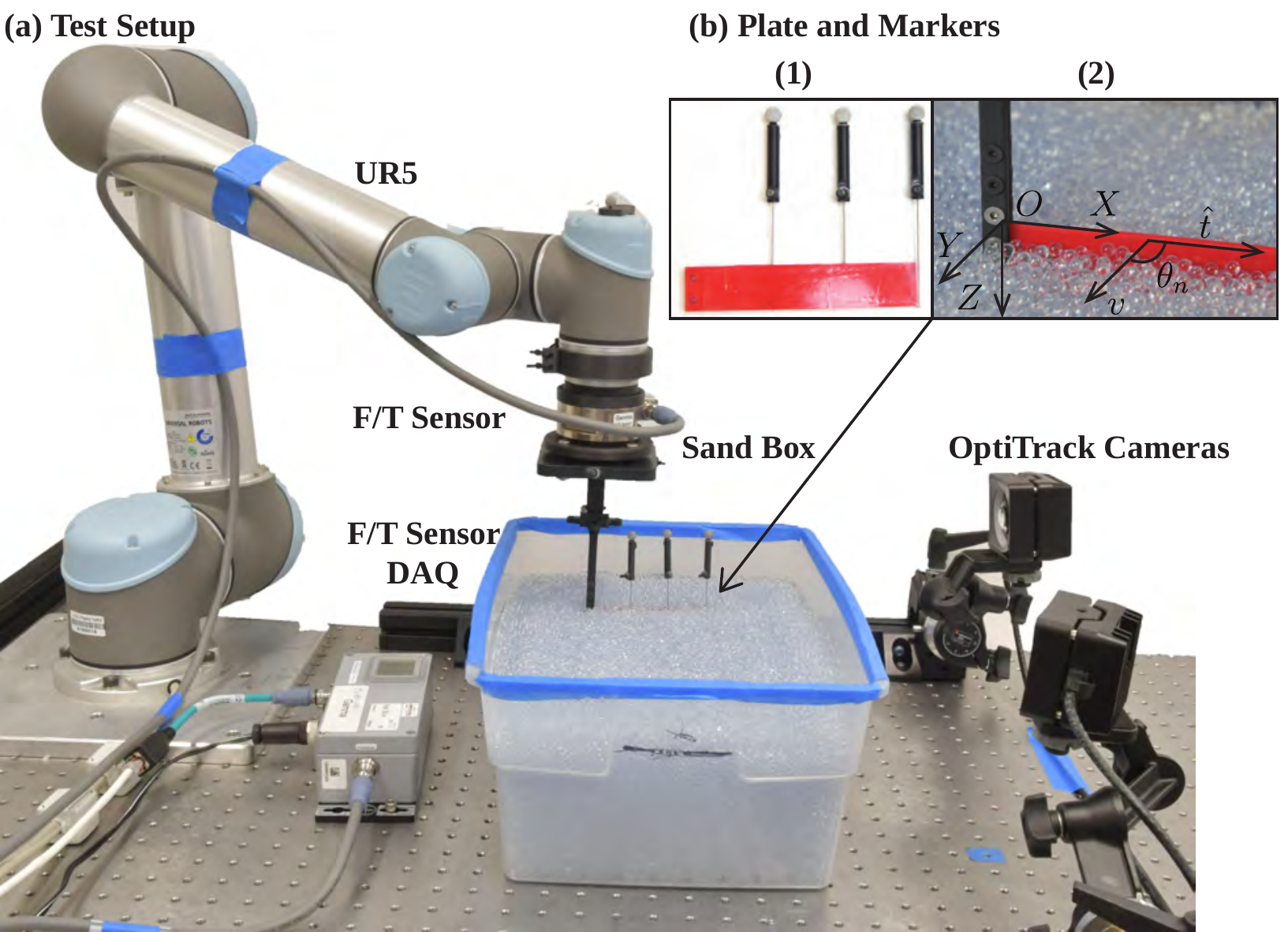}
\caption{\textbf{The test setup.} {(a)} shows the test setup used in this study, including a UR5 robot arm, an ATI Gamma Force/Torque sensor with DAQ, and a group of OptiTrack cameras. The plate is attached to the test setup using a custom 3D-printed attachment and inserted into the box placed on the test table. OptiTrack markers are connected to tiny steel rod that mounted to the plate. The cameras are moved for pictures, {(b)}: A closer look of the plate.
In (b-1), the markers are placed on the plate using the extension rods and (b-2) illustrates how the plate is inserted into the box and the local coordinate of the plate.}
\label{fig:reciprocal_force}
\end{figure}

We use the test setup illustrated in Fig. \ref{fig:reciprocal_force} to measure force, control displacement and track markers' position in experiments. The mounting side of a 6-DOF force/torque sensor (ATI Gamma F/T Sensor) is connected to the end of a robot arm (Universal robot, UR5). The intruders and fins are mounted to the tool side of the sensor with different custom attachments in various tests; these are inserted into a box filled with glass beads. Currently, the average particle diameter, $d_g$, is 4mm. The robot arm is programmed to hold, drag or rotate the object using a Python script, which also records the markers' locations over time using a motion-capturing camera system (OptiTrack).
To track the deformation inside the glass beads, we attach the OptiTrack markers using extension rods (diameter = 1\,mm) along each joint axis so that they can be seen above the beads, as in Fig. \ref{fig:reciprocal_force} (b). Four markers are placed on the plate and on the center of UR5 end effector, aligning with the origin of the plates. The markers attached to the intruder are mounted so as to obtain its location, curvature and displacement\dl{.  The maximum error introduced on force estimates due to twisting in the $YZ$-plane is less than 0.3\%, which is thus ignored.}

\subsection{Granular RFT for Compliant Plates}
\label{carft}
\begin{figure}[h]
\centering
\includegraphics[width =0.85\columnwidth]{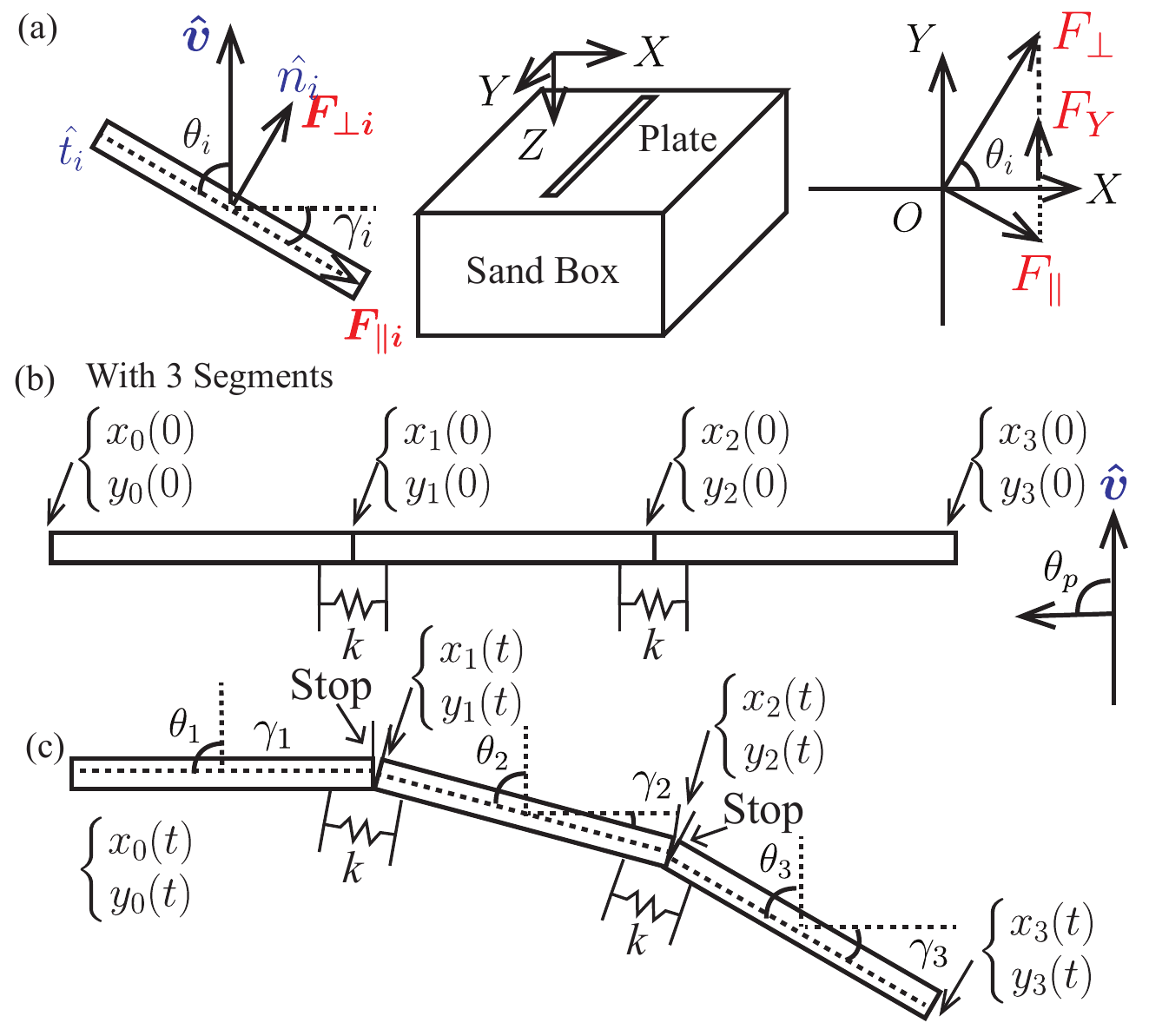}
\caption{
\textbf{How the granular RFT can be applied to the compliant plates.}
(a) On the left side, we show a single segment from {(c)}, $v$ is the direction of the velocity, \dl{$\bm{\hat{n_i}}$, $\bm{\hat{t_i}}$} indicate the normal and tangential direction of the plate, \dl{ $\theta_i$} is the angle of attack of \dl{\textit{i}-th} segment of the plate and \dl{$\gamma_i$} is the angle of twist of each segment, where \dl{$\theta_i + \gamma_i = 90^{\circ}$}; the other two represents the local coordinate of the test setup and the propulsive/resistive force, $F_Y$.
(b) An undeformed plate with 3 segments at $t=0$, the equations under the plate show the position of each point along the hinge.
\dl{On the right side of (b), we show the local frame of the plate.}
(c) The deformed, final geometry of this origami inspired plate \dl{at later time $t_{f}$. In (b) and (c), we also showed how equivalent springs and stops were placed on the plate.}
}
\label{fig:rft_plot}
\end{figure}

To motivate the problem, consider a multi-link intruder moving within a frictional, granular environment as seen in Fig \ref{fig:rft_plot}. As it is dragged through the medium it bends and deforms through its interactions with the glass beads. If starting from rest in its undeformed flat state, the final configuration of a compliant intruder will adapt to achieve a minimum energy configuration, impacting the forces it imparts on the media; in contrast, a rigid plate, dragged through the soil, will see far smaller deflections \dl{and higher forces}; its effective configuration after being dragged remains unchanged, and its force interactions can be completely described by traditional RFT. The goal of our modeling approach is to consider the effect of compliance, which will require augmenting current RFT theory. This has been considered previously by Peng et al \cite{peng_maximizing_2017,peng_propulsion_2017}, but applied to problems without changing stiffness in the gait cycle, such as a rigid slender body with torsional spring at the origin or a flexible filament that can bends in both directions.
A compliance-augmented granular RFT model will provide a more convenient way to model flexible materials interacting with granular material, and will permit more rapid analysis and optimization of compliant sand swimmers and diggers. We propose to utilize this theory to model our fins and better understand robot locomotion in granular media.

To demonstrate how compliance could be applied to RFT on our origami inspired plate ,we first consider a three-link plate, seen in Fig. \ref{carft}, where the linkages are equally spaced along the long edge of the plate.
The plate, with height, $L_p$, linkage number $N=3$ and joint stiffness/equivalent spring constant at hinge, $k$, is placed vertically in the \textit{xy}-plane and dragged inside the bulk material at a speed $v_p$ with angle of attack, $\theta _p$, as shown in Fig. \ref{fig:rft_plot}{(a)}. The normal and tangential vector of \dl{plate $i$ are $\bm{\hat{n}_i}$ and $\bm{\hat{t}_i}$} respectively; the angle of attack of the plate can thus be written as \dl{
$\theta_p = \angle (\bm{\hat{t}}, \bm{\hat{v}})$}.
Based on previous work focusing on the effect of intruder speed on dragging force in granular material, we assume that the resistive force on the plate, $F_p$ is insensitive to speed at low, friction-dominating speeds \cite{ wieghardt_experiments_1975, PhysRevE.64.061303, albert_slow_1999, maladen_undulatory_2009}.
In traditional granular RFT, the resistive force in the moving direction on a equivalent rigid plate ($N=0$) moving at steady state is calculated using
\begin{equation}
\mathbf{F_p} = \int_{0}^{L}({ {d \mathbf{F_\perp} sin(\theta_p)}-d \mathbf{F_\parallel} cos(\theta_p)})
\end{equation}
where $dF_{\parallel}$ and $dF_{\perp}$ are the force derivatives per element.

We now consider the case for a three-link origami-inspired compliant plate, where the length of each segment $l$ equals $\frac{L_p}{N}, (N=3)$; \dl{$\theta_i$} stands for the angle of attack for each segment. In granular RFT, force on individual segments are only influenced by the field of the local granular material; we can thus assume that for all segments, \dl{$dF_{\parallel i}$ and $dF_{\perp i}$} remain the same, given that the granular media remains in a quasi-static state. 

The instantaneous position of the joint on plate $i$ \dl{can be seen in Fig. \ref{fig:rft_plot}, as indicated by $x_i(t)$ and $y_i(t)$.}
The force acting on each joint can be expressed as \dl{$F_{Yi}$},
\begin{gather}
\dl{
F_{Y_i} = \int_{0}^{\frac{L}{N}}({ {d \mathbf{F_\perp}sin(\theta_i)}-d \mathbf{F_\parallel} cos(\theta_i)}) 
}
\end{gather}
Note that when \dl{$i=1$}, $\theta_{0}$ refers to $\theta_p$, which is the angle of attack about the axis located at the origin. In this case, after sufficient time, the intruder traveling at constant velocity $v$ reaches a final state at time $t_f$, at which point the resistive force and geometry of the plate remains steady.
For instance, when the plate is placed vertically in glass beads and dragged at a constant speed $v$ at angle of attack $90^{\circ}$, the boundary conditions will be as follows:
\dl{
\begin{equation}
\left\{\begin{matrix}
x_{0}(0) = 0 \\
y_{0}(t) = vt \\
x_{N}(0) = L_p \\
y_{N}(0) = 0 \\
\theta_{p}(t) = 90^{\circ}\\
\end{matrix}\right.\\
\label{bcs}
\end{equation}
}
The decomposed segments and connection between the plate are connected to our test setup, as shown in Fig. \ref{fig:rft_plot}. We may calculate the steady state configuration of the system at $t_f$ by tracking the position of the markers attached to the plate, which permits us to track the position and deformation of the compliant intruder. 
This configuration corresponds to the steady-state force-balance between the internal compliance of the fin and the external forces exerted by the granular media as it moves.
\section{Material and Methods}
\subsection{Compliant Fins for Locomotion}
\label{section2b}
To understand how changing the stiffness of intruders can change the resistive force interactions, we propose three different fin designs including rigid, fully-compliant and origami-inspired plates; we hypothesize that the origami-inspired plate with a one-sided stop can behave differently in different portions of a gait cycle as opposed to the fully rigid and fully compliant plates.

To change the stiffness of the plates in a full gait cycle, we add stop on one side of the fin to permit the flexible sheet/joint to bend only during the recovery portion of a flapping gait.
When the origin of the plate oscillates back and forth in the granular media, the fin only bends in the recovery-stroke, recovers to flat while approaching the constraint and remains rigid against the stop during the power stroke, effectively behaving as a rigid plate in one direction and a compliant plate in the other.
Our intent is to use the difference in thrust force between power and recovery portions of its stroke to generate forward motion. Here we present the plate with unidirectional bending by embedding a joint limit.

\begin{figure}[h]
\centering
\includegraphics[width =1\columnwidth]{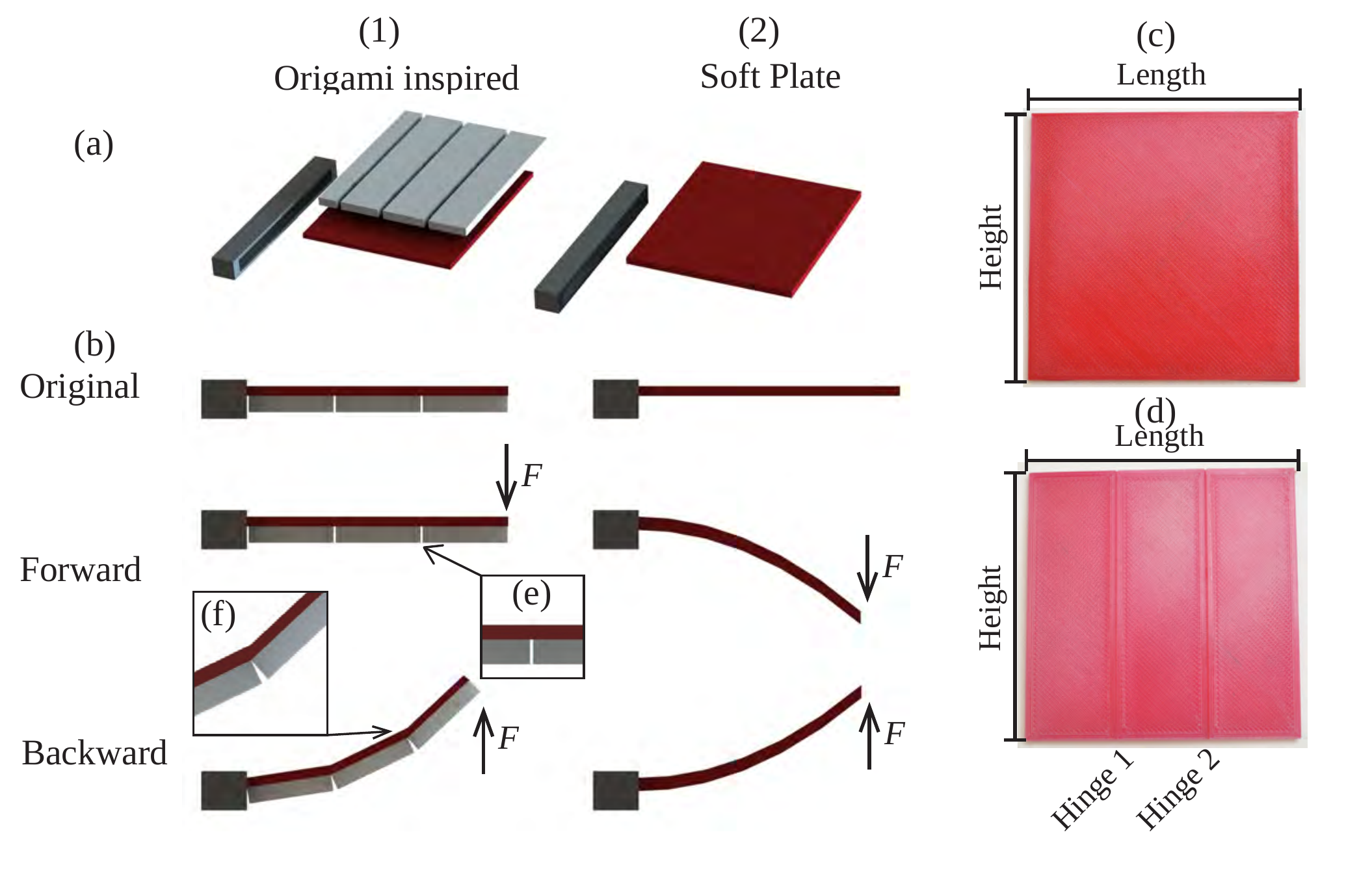}
\caption{\textbf{Fabrication of the fins.} Column (1) is the origami plate while (2) is the soft plate.
(a) Assembling of the parts in the fin: gray material stands for frame, printed by conventional PLA filament; red layers are printed with red Ultimaker TPU 95A as flexible sheet and white plates are made with Ultimaker transparent nylon.
(b) The bending of the actuator plates and effect of the stops; in (b-1), the nylon acts as the rigid stop.
(b-2) The normal flexible plate has no stop mechanism thus can bend in two directions. Due to the accuracy and line width of our printer, we draw the hinge in our 3D design with a gap of 0.4mm. When finished, the printed plates have a gap of 0.05mm and do not permit bending beyond the nylon surface.
(c) Printed soft plate.
(d) Printed origami plate with two hinges.
\dl{(e) and (f) Enlarged view for the origami hinge and rigid stop in normal and bending states respectively.}
}
\label{fig:actuators}
\end{figure}

The plates of the fins, which can be seen in Fig. \ref{fig:actuators}, are printed using Ultimaker TPU 95A with Ultimaker Nylon, in comparison with the fully soft plate design that only uses Ultimaker TPU95 soft filament.
The nylon, when printed on top of the flexible TPU layer, bonds firmly to the TPU and serves as a rigid layer.
A 0.05 mm gap separates rigid links, exposing only a small portion of the TPU to bending in one direction. In the other direction, the narrow gap quickly closes, causing interference between neighboring Nylon links, behaving effectively as a joint limit.
\dl{In Fig. \ref{fig:rft_plot} (b) and (c), we present conceptual diagrams of the directional bending and stop.}
Fig. \ref{fig:actuators} highlights the geometry and behavior of this plate in comparison to a flexible plate.
\dl{We replicated the origami/laminate structure \cite{onal_towards_2011, 6266749, wood_first_2008,tolley_self-folding_2014} using a 3D-printing approach, where the nylon serves as to rigidze the sheet and TPU serves as the flexure joint. The gap between nylon links permits bending in one direction while acting as a joint limit in the other. The nylon and TPU 95A bond during the printing process, eliminating the need for an explicit adhesive layer.
A detailed discussion of the modelling approach can be found in Section \ref{Pynamics}.}
\subsection{Model Verification for Compliant Fins}
\label{Modelverify}
To verify our model, a rigid plate is mounted vertically to our test setup.
We first actuate the robot arm with a triangle wave input to drag the plate back and forth a distance $D$ = 200 mm with a velocity $v$ = 30 mm/s at angle of attack, $\theta_p$ = $90^{\circ}$. The plate is then rotated by the UR5 at different angles, $\Theta$ = $120^{\circ}$ with angular velocity, $\omega$ = 60$^{\circ}$/s. The resistive forces/torques vs velocities are thus obtained across different sets of actuation conditions of this group of experiments.
In this group of rigid plate tests only, the plate is dragged at a constant velocity while its angle of attack is changed to obtain the RFT parameters, $dF_\parallel$ and $dF_\perp$, which can supply the dynamics simulation with experimental data.

In each tests the plate is oscillated five times. During the first sweep, the plate is observed to sweep away and excavate some granular material, resulting in an inconsistent force measurement. Subsequent oscillations do not further excavate any material, resulting in a more consistent force curve; we thus have removed the first cycle of data from our results. The plot in Fig. \ref{fig:force_result} shows the resulting force pattern of the second to fifth cycle.

A set of soft/origami plates with different stiffness, $k$, is then mounted to our test setup using the same dragging or rotating input to obtain the behavior of a compliant intruder. We can now compare the performance of each plate and discuss the result in the next section.
We observe that the resistive force of a compliant plate decreases as it deforms, resulting in a smaller drag force than the equivalent rigid plate.

\section{Result and Analysis}
\subsection{Compliant Plate Dragging and Rotating }
\label{actuator_experiments}
\begin{figure*}[h]
\centering
\includegraphics[width =1.9\columnwidth]{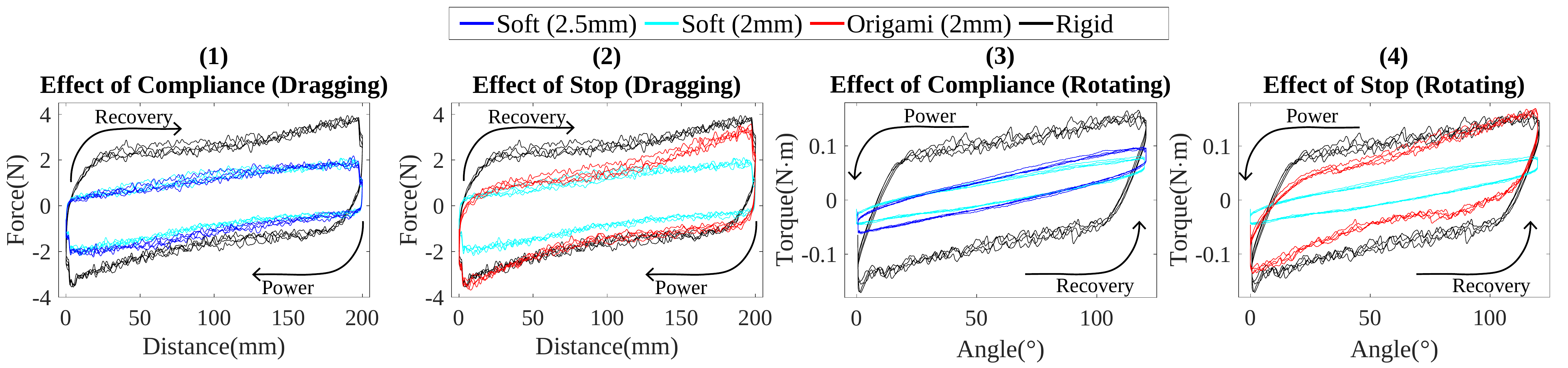}
\caption{
\textbf{Dragging and rotating results.}
The legend on the top indicates the plate type.
These sub-figures show how the force/torque change corresponding to the distance/angle. For example, in (1) \dl{and (2)}, the plate moves from 0 mm to 200 mm as recovery stroke (Upper and left) then is dragged back to the origin in the power stroke (Lower and right).
(1) and (2) show the dragging results, \dl{where rigid and soft (2mm) are shown in both (1) and (2). In contrast, soft (2.5mm) is only presented in (1) and origami (2mm) can be seen in(2). Following the same principles,} (3) and (4) indicate the rotating results. 
}
\label{fig:force_result}
\end{figure*}

\begin{figure}[h]
\centering
\includegraphics[width = 0.9\columnwidth]{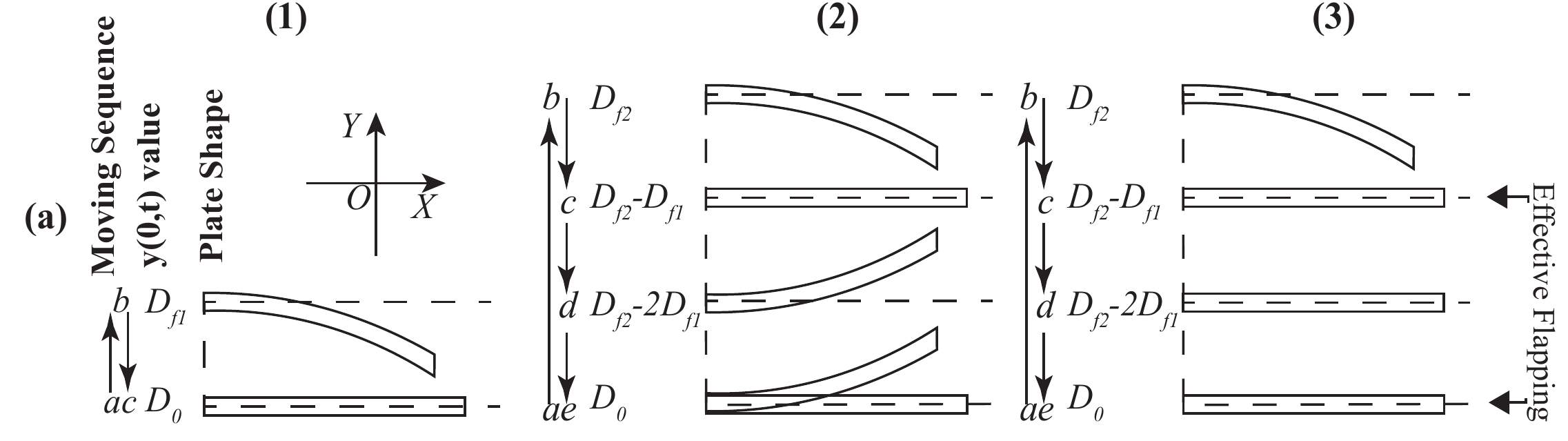}
\includegraphics[width = 0.9\columnwidth]{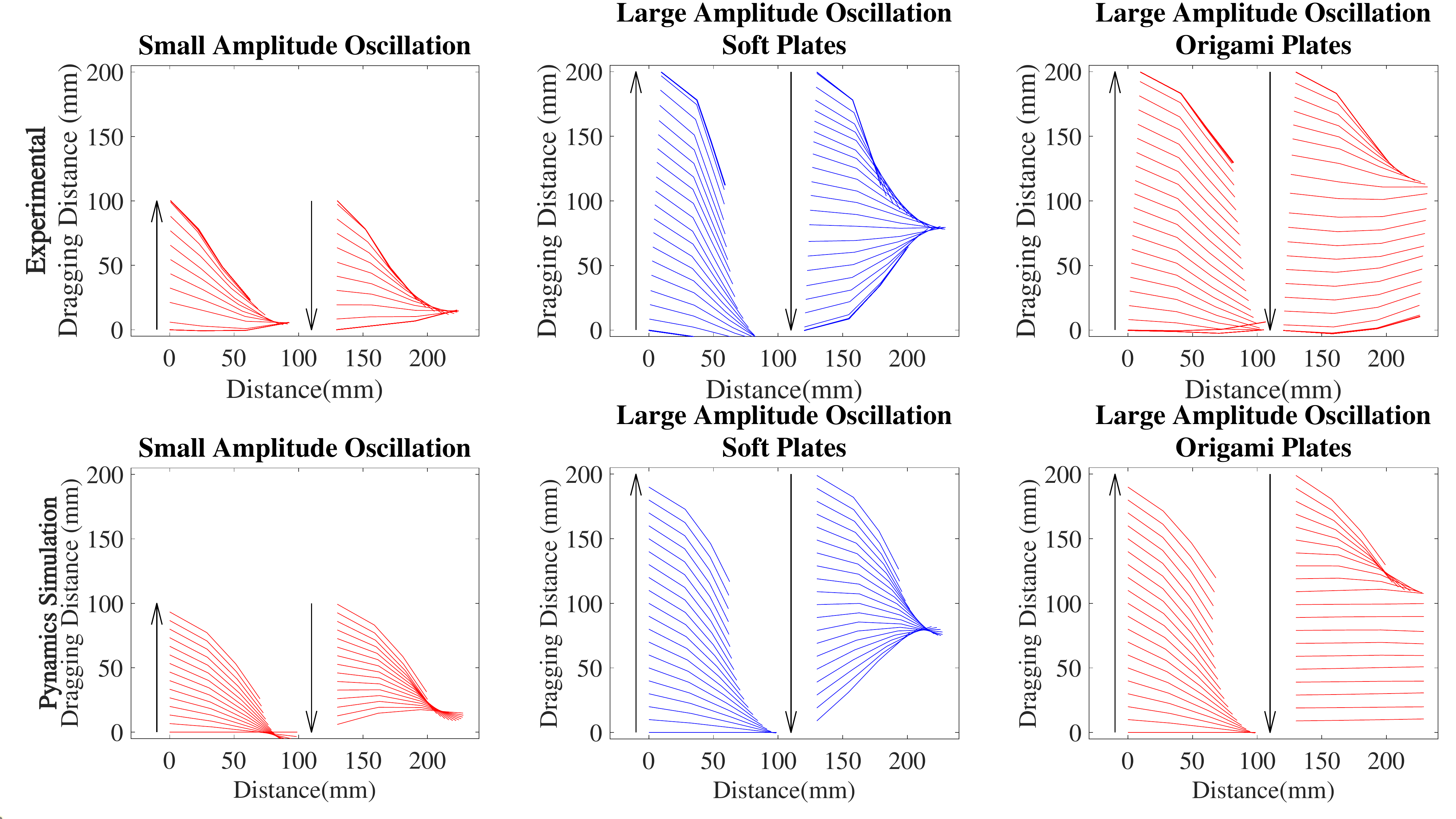}
\caption{
\dl{
\textbf{Concepts of effective flapping.}
Column (1): Small amplitude oscillations; (2) Large amplitude oscillation for flexible plate. (a-3) Large amplitude oscillation for the origami plate.
Row (a) shows a schematic diagram of the bending of the plate. Letters, arrows and notations on the left side of each sub-figure indicates the moving sequence and distances (value of $y(0,t)$), following alphabetical order; we plot the local coordinate frame on the upper-right corner of (a-1).
In row (b) and (c), we show the bending process of plate dragging tests corresponding to row (a) from experimental data and simulations using Pynamics, respectively. The legends indicate the plate types.
In each sub-figure, the arrows in row (b) and (c) show the dragging directions. And on the left side, we show the power strokes and the right hand sides are the recovery strokes.
}
}
\label{fig:effective_flapping_dragging}
\end{figure}
Though our robot's proposed actuation is based on rotation, as seen in Fig. \ref{fig:robot}, we believe that dragging tests and results may be extended to rotation cases and can help us to gain insights into the contribution of compliant plates as well as compare the merits of different designs.
The force vs. distance and torque vs. angle for linear and rotational cases, respectively, are plotted in Fig. \ref{fig:force_result}. 
Compared to a rigid plate, the force, $F_p$ for soft plates (blue and cyan) is lower due to a smaller effective profile in the direction of motion when deformed in the forward and reverse directions. A symmetric pattern is observed for rigid and soft plates.
The origami plate, with a one-sided rigid stop and compliance in the other direction, shows a different pattern in contrast. As seen in Fig. \ref{fig:force_result}(2), the origami-inspired plate in red produces an asymmetric force-displacement histogram in its power and recovery stroke. We observe that, by introducing stops, the new curve overlaps with the forces exerted on the rigid plate (in black) in the power stroke, and much more closely tracks the forces exerted on the soft plate (in blue) in the initial portion of the recovery stroke. The force near the end of the recovery stroke slowly deviates toward the rigid limit, which we attribute to a higher equivalent stiffness compared to the soft plate. 
In the case of rotation, we find similar results. In Fig. \ref{fig:force_result}(4), compared to the soft plate, the torque generated by the origami-inspired plate can effectively act as a rigid plate in the power stroke. In the recovery stroke, we find the torque of the origami-inspired plate is higher than the compliant continuum plate in Fig. \ref{fig:force_result}(3), due to the differences in the thickness and equivalent stiffness between the origami plate and fully compliant plate.
We may use this finding to design our robot by integrating joint limits into the fin design, thus breaking symmetry.
\subsection{Effective Flapping for Plates}
\label{effective_flapping}
When an origami plate is dragged inside a granular medium from origin $D_0$, the plate will reach a maximum deflection as the dragging distance reaches $D_{f1}$.
Once the maximum deflection is reached, the resistive forces of the plate remain constant. A displacement in the opposite direction will drive the plate towards its original shape, where the threshold value required for the plate to return its original shape is $D_{f1}^{r}$. If the plate doesn't start to bend in the opposite direction as it returns to its origin, we refer to this as ``Small amplitude oscillation", as seen in Fig. \ref{fig:effective_flapping_dragging}(a-1). However, if the plate's amplitude increases and a change of the direction occurs at $D_{f2}$ ( $D_{f2} > D_{f1}$) when oscillating back, the plate will recover to a flat shape at $D_{f2}-D_{f1}^{r}$.
At a constant speed the plate's configuration remains constant after $-D_{f1}^{r}$ when traveling back to the origin; we refer this scenario as ``Large amplitude oscillation", as shown in Fig. \ref{fig:effective_flapping_dragging}{(a-2)}. Under large amplitude oscillation, if a unidirectional joint limit is established for $\gamma_i<0$ (angles of twist in Fig. \ref{fig:rft_plot}), the joint limit will break motion symmetry, and non-zero total thrust forces will be produced over a full power/recovery stroke during $(D_{f2}-D_{f1}^{r})$ to $D_{0}$, as in the right side of Fig. \ref{fig:effective_flapping_dragging}{(a-3)}. We refer this phenomenon as ``Effective Flapping''. Thus in the design of the fin, an important guideline is to ensure that the amplitude of the fin is larger than $D_{f1}$ so as to guarantee nonzero net force. In Fig. \ref{fig:effective_flapping_dragging}{(b-3)}, we observe that, when dragged back, the markers' locations indicate that a small amount of bending occurs when $\gamma_i<0$; this can be attributed to a number of potential reasons: (1) Error between the marker attached to the end of the extension rod; (2) a non-zero joint limit due to geometric differences between the ideal and prototyped plate; or (3) because the rigid material's stiffness plays a role in establishing a high -- but not infinite -- joint stiffness at the joint. In the following section, we will discuss the simulation result using \dl{Pynamics}.

\subsection{Simulation of Plate Dragging}
\label{Pynamics}
To simulate the plate's motion, we first define a 3-link origami-inspired plate with the extension rod as a 4-link serial chain of links connected by pin joints, where the first linkage is the extension rod and the next three links correspond to the segments of the origami-inspired plate (seen in the supplemental video). The forces and torques applied to each individual plate include the forces due to the granular RFT model ($dF_{\perp}, dF_{\parallel}$) applied to the geometric center, and an equivalent torsional spring applied to each joint, representing material-based compliance as a linear constant $k$ in the soft direction.
These values are obtained by experimental measurement using our test setup. To model the joint limit's stiffness when $\gamma_i<0$, we apply an arbitrarily large spring constant, $k_r =10000$\,N/rad. Also added to the model is a rotational damper located at each joint, whose damping ratio, $b$, is used to model the loss within the soft material itself. Other dynamic parameters include mass and \dl{inertia}, which are obtained based on the density and geometry of the materials used \dl{and essential for solving for acceleration prior to integration in Pynamics}.

With the parameters described above, one can obtain the time-based motion of the system in our dynamic model as a function of specific motion or force inputs. To replicate the motion data obtained experimentally for the origami-inspired plate, we supply a motion constraint for $v$, the forward dragging velocity. We supply the previously obtained experimental values for $dF_{\parallel}$, $dF_{\perp}$ and $k$ as system constants, while the value for $b$ is solved iteratively using the Covariance Matrix Adaptation Evolution Strategy (CMA-ES), which is provided in Python via the pycma \cite{hansen2019pycma} package.
In order to do this, $b$ and $v$ are supplied as inputs to a function for calculating the angles of twist of each joint ($\gamma_1$, $\gamma_2$, $\gamma_3$) through time, $t$, as:
\begin{equation}
 F_{Sim}(b,v,t) = [\gamma_{s1}, \gamma_{s2}, \gamma_{s3}]^{T}
\end{equation}
In comparison, we can thus write the experimental system's state as:
\begin{equation}
 F_{Exp}(v,t) = [\gamma_1, \gamma_2, \gamma_3]^{T}
\end{equation}
To obtain the joint damping ratio, $b$, we first calculate the Mean Squared Error (MSE) for the joint angles over time as an error function:
\begin{equation}
 E(b) = \sum_{v=10}^{50} \frac{( F_{Sim}(b,v,t) - F_{Exp}(v,t))^2}{5}
\end{equation}
We then solve $E(b)$ using pycma to find the value of $b$ via:
\begin{equation}
 min \;\; E(b)
\end{equation}
\begin{figure}[h]
\centering
\includegraphics[width =0.9\columnwidth]{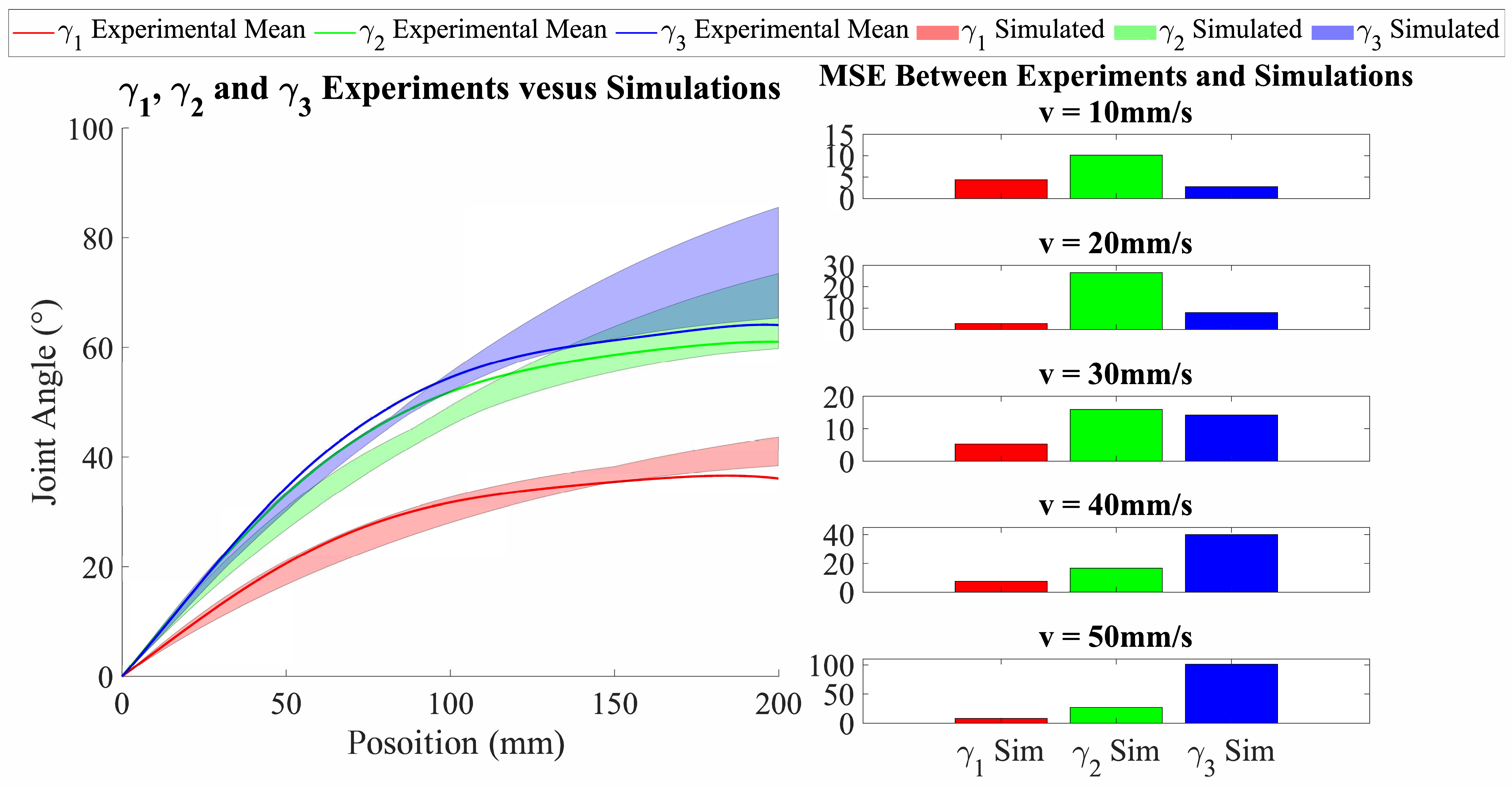}
\caption{
\textbf{Simulation result of plate deformation using Pynamics}
In {(a)}, we show the joint angles over distance comparing to the simulation result. Solid lines show the mean of the experimental values of $\gamma_1$, $\gamma_2$ and $\gamma_3$. \dl{The transparent region shows the range of simulation results across five speeds (10mm/s - 50mm/s), where the height of the variation stands for the minimum and maximum simulated values for each position in mm.}
Column {(b)} indicates the MSE of each speed between the simulation and experiments. Each row inside is the MSE between the simulation result and the experimental data at the same velocity. Red, green and blue bars stand for $\gamma_1$, $\gamma_2$ and $\gamma_3$, respectively.
}
\label{fig:sim_error}
\end{figure}

In order to evaluate the error of the simulation against our experimental results, the experimental joint angles $(\gamma_{1}, \gamma_{2}, \gamma_{3})$ are extracted from experimental motion data across five specified velocities (10-50 mm/s) in the recovery stroke from flat (drag the plate in the soft direction); these are shown as solid lines in Fig. \ref{fig:sim_error}(a). These data are then compared against the simulated trajectories ($\gamma_{s1}$, $\gamma_{s2}$, $\gamma_{s3}$), which are displayed as the multicolored regions in Fig. \ref{fig:sim_error}\dl{, represent the maximum errors of each joint angle against the experimental mean value.}
In the experiment, joint angles are primarily affected by the configuration-dependent loading conditions as the system is dragged through the granular media.
As a function of the base position, the nominal speed of the test plays little role in the configuration of the system, in agreement with RFT; we thus show the experimental data in Fig. \ref{fig:sim_error}(a) as the mean across all speeds.
The MSE across different speeds is shown separately as well, as seen in Fig. \ref{fig:sim_error}(b).

The results show that with a single damping parameter $b$, obtained by finding the value that minimizes the error between simulation and experimental results, the \dl{Pynamics}-based dynamic model is able to predict the motion of the origami-inspired plate over time. From these results we see that \dl{Pynamics} over-estimates the deformation of the distal-most joint and underestimates the early motion of all joints. This can be attributed to our use of a velocity-based damping model for lumping together all dynamic material effects around each joint; a more refined model of the material-based loss within our plate could capture other effects, at the risk of over-fitting. However, these results demonstrate that the error between model and experiment are reasonable enough to use the model to understand and contrast thrust generation across various gaits.
\subsection{Robot Design and Tests}
\label{Robot_tests}
\begin{figure}[h]
\centering
\includegraphics[width = 0.9\columnwidth]{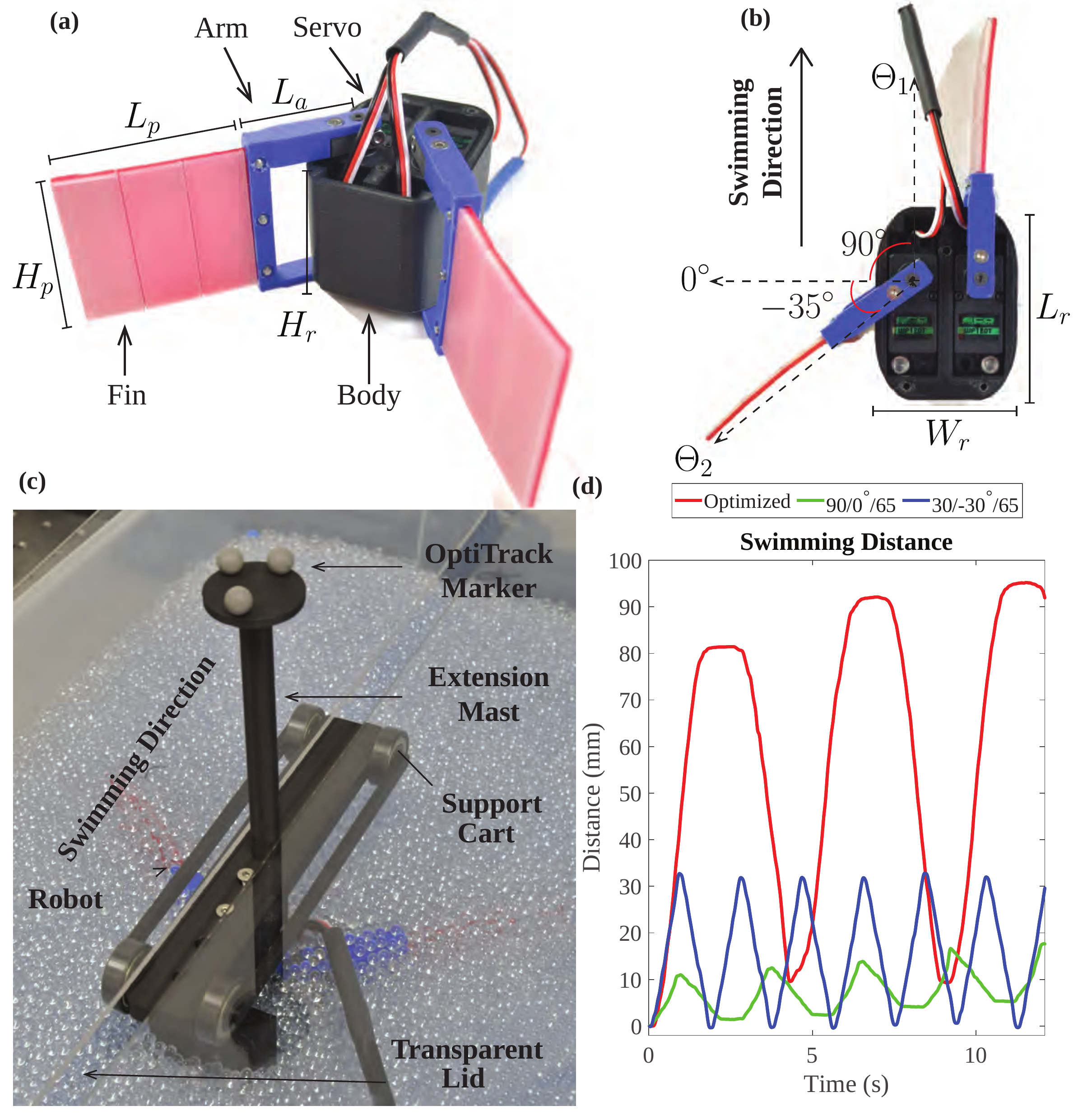}
\caption{\textbf{The granular swimming robot prototype, other details and the swimming results.}
{(a)} and {(b)} are the perspective and top of the robot, respectively. We also include the design parameter associated with this robot.
{(c)} shows how the robot is placed in the glass beads with a a support cart with extension mast with markers on the top to track location.
{(d)} illustrates the trajectories of the robot in the moving direction shown in {(c)} over time in tests for different configurations. Legends indicates the value of $\Theta_1 / \Theta_2$ / $L_a$, respectively. For instance, 30/-30$^{\circ}$/65 stands for experimental trajectory of configuration, $\Theta_1 = 30^{\circ}$, $\Theta_2 = -30^{\circ}$ and $L_a = 65 mm$.
}
\label{fig:robot}
\end{figure}

To determine if the plates can be used in the generation of motion, we introduce a full robot model consisting of a body and two origami-inspired fins.
We adapt the basic mechanical design of our robot from the sand-burrowing robot in \cite{russell_crabot_2011}. Our robot, seen in {Fig. \ref{fig:robot}{(a)} and {(b)}}, consists of a 3D-printed case and two bilaterally-symmetric sets of servos and fins, attached to the main body of mass, $M_r$, height, $H_r$, width, $W_r$, and length, $L_r$. The servo angles are represented by $\Theta_1$ and $\Theta_2$, and the resistive force in the moving direction is represented by $F_r$. The robot servos are connected and controlled by an Arduino Uno using PWM signals sent from an attached computer.

To select a set of feasible parameters of the robot, we first consider that the resistive force for the robot body should be minimized.
In granular RFT, resistive force increases with the cross section; we thus select the robot body with the following parameters: $H_r$ = 45\,mm, $W_r$ = 55\,mm, $L_r$ = 75\,mm and $M_r$ = 220\,g. These are the minimum sizes and corresponding masses we can design and manufacture to accommodate the servos.
Constrained by the size of robot body, we select the length of the plate, $L_p$ of 60\,mm and the height of the plate, $H_p$ as 55\,mm to accommodate the robot arm and actuator frame.
\dl{We define the minimum value of the length of the robot arm, $L_a$ as 60\,mm, which is the lowest value we could select to allow the full range of motion.
}
After selecting these parameters, we mount the plates to the robot then mount the system to our test setup, inserting it into the box in preparation for the next step. 

To utilize RFT-based deformation found in Section \ref{actuator_experiments} and Section \ref{effective_flapping}, we obtain a new set of $dF_{\perp}$ and $dF_{\parallel}$ values. This is due to differences in the sizes and depth of the plates from the previous section.

When the robot is on, servos flap \dl{the fins of length $L_a$} between $\Theta_1$ and $\Theta_2$.
We define the power stroke as the portion of the stroke when the fins rotate from $\Theta_1$ to $\Theta_2$; the remainder of the stroke is defined as the recovery stroke. During the power stroke, the robot moves forward for $D_1$\,(mm), while during recovery, the robot moves backward by $D_2$\,(mm). To measure the efficiency of the robot, we define the swimming efficiency as:
\begin{equation}
\dl{
\eta (\Theta_1,\Theta_2) = \frac{D_1 - D_2}{D_1}
}
\end{equation}
\dl{which represents the proportion of the net displacement, $D_1 - D_2$ in the forward distance, $D_1$ of the power stroke.
As discussed in
Section \ref{Pynamics} and Fig. \ref{fig:sim_error}, in RFT-regime, force and deformation of the compliant fin are configuration-dependant rather than velocity-dependent; we conclude that increasing the angular velocity of the inputs will not change the $\eta$ significantly. Faster-oscillating fins will simply drive the robot at a higher speed.  Thus we have selected efficiency as the performance metric.
}

An optimization strategy was used to optimize the efficiency $\eta$ by tuning $\Theta_1$, $\Theta_2$ \dl{and $L_a$}. In a symmetric gait both sides of the robot are actuated symmetrically; we thus simulated the right half of the robot to increase the optimization speed. The objective function was selected as:
\begin{equation}
 min \;\; (1-\eta(\Theta_1,\Theta_2, L_a))
\end{equation}

\dl{With a minimum value of $L_a$ as 60 mm, the servos and fins have a maximum ranges of motion of $0^\circ \sim  90^\circ$. We thus define the bounds of the optimization of $\eta$ as $\Theta_1 \in [0^{\circ},90^{\circ}]$, $\Theta_2 \in [-90^{\circ}, 0^{\circ}]$ and $L_a \in$ [60mm, 85mm], where the upper bound of $L_a$ is the longest value we can manufacture.}
The optimizer found that \dl{$\Theta_1$ = $60^{\circ}$, $\Theta_2$ = $-90^{\circ}$ and $L_a$ = 65mm} produces the maximum swimming efficiency of \dl{$\eta_{best}$ = $8.95\%$.}

\begin{table}[h]
\begin{center}
\begin{adjustbox}{width={\columnwidth},totalheight={\columnwidth},keepaspectratio}%
\begin{tabular}{c|ccccccc}
\hline
\textbf{Configuration}
& \begin{tabular}[c]{@{}c@{}}$\mathbf{\eta_{Exp}}$\\ (\%)\end{tabular}
& \begin{tabular}[c]{@{}c@{}}$\mathbf{\eta_{Sim}}$\\ (\%)\end{tabular}
& $\mathbf{MSE}$
& \begin{tabular}[c]{@{}c@{}}$\Theta_1$\\ ($^{\circ}$) \end{tabular}
& \begin{tabular}[c]{@{}c@{}}$\Theta_2$\\ ($^{\circ}$) \end{tabular}
& \begin{tabular}[c]{@{}c@{}} \dl{$L_a$}\\ \dl{(mm)} \end{tabular}
& \begin{tabular}[c]{@{}c@{}} \dl{\textbf{Net Speed}}\\ \dl{(mm/s)} \end{tabular}
\\ \hline
Optimized   & \dl{8.61} & \dl{8.95} & \dl{$1.2 \times 10^{-5}$} & \dl{60}   & \dl{-90}   & \dl{65}    & \dl{1.69} \\ \hline
Symmetry 1  & \dl{1.22} & \dl{0.06} & \dl{$1.35 \times 10^{-6}$} & \dl{30}   & \dl{-30}   & \dl{65}   & \dl{0.48} \\ \hline
Symmetry 2  & \dl{1.08} & \dl{0.05} & \dl{$1.06 \times 10^{-6}$} & \dl{30}   & \dl{-30}   & \dl{75}   & \dl{0.59} \\ \hline
Asymmetry 1 & \dl{3.12} & \dl{3.71} & \dl{$3.5 \times 10^{-5}$} & \dl{90}   & \dl{0}   & \dl{65}   & \dl{0.34} \\ \hline
Asymmetry 2 & \dl{3.57} & \dl{3.65} & \dl{$6.4 \times 10^{-7}$} & \dl{90}   & \dl{0}   & \dl{75}   & \dl{0.43} \\ \hline
\dl{Rigid Fin} & \dl{0.11} & \dl{1.02} & \dl{$8.281 \times 10^{-5}$} & \dl{60} & \dl{-90} & \dl{65} & \dl{0.02} \\ \hline
\end{tabular}
\end{adjustbox}
\caption{ \textbf{Robot Swimming Efficiency versus Configuration.}
}
\label{robot_eff}
\end{center}
\end{table}

To measure the swimming efficiency of the flapper robot, we submerge the robot 1\,cm deep into the container filled with 4\,mm glass beads before each run. During initial experiments, we observe that the robot rolls and rises above the surface of the granular media due to drag-induced lift \cite{ding_drag_2011}.
\dl{This has been addressed in related work by using a wedge attached to the head of a sand-swimming robot  \cite{maladen_granular_2011}. However, given the small size and minimal design of our robot, adding such a similar wedge would introduce new
\dma{complexity to the robot that does not directly answer our main research question. To simplify our prototype and experiments}}
we place a lid on the box and mount the robot to a supporting cart, whose wheels remain in  contact with the lid  to constrain the robot in a constant horizontal plane at a fixed depth under the granular media. To track the location of the robot under the lid cover, an extension mast with OptiTrack markers is attached to the cart (weight=160\,g), permitting the motion of the robot to continue to be tracked under the beads within the box, as seen in {Fig. \ref{fig:robot}{(c)}}.
We have compared the swimming efficiency of various configurations of the robot's joint limits in \dl{Table \ref{robot_eff}} and verified the optimized swimming efficiency from \dl{Pynamics} experimentally.
Our dynamic model accurately predicts the swimming efficiency of the robot; however, it over-predicts the forward and backward distance covered. This is due to the fact that our current granular RFT model over-estimates the resistive forces and fin deformation, resulting in higher distances covered in both directions. The trajectories of each experiment can be seen in {Fig. \ref{fig:robot}{(d)}}, illustrating the difference in efficiency as a function of gait.

\section{Conclusion and Future Work}
We have presented an origami-inspired compliant fin for granular locomotion using a new approach for augmenting compliance into a traditional granular RFT model. This has been applied to the model for a two-fin robot and compared to the experimental prototype. We have shown that the time-based evolution of bending and recovering of the fins, modeled by material damping within the fin can be used to improve the swimming performance of a robot.

Our results have detailed an approach to understand the principles by which nonlinear compliant materials can be leveraged within granular media, providing potential design simplifications that can reduce control overhead in the future. Future work includes using the compliance-augmented granular RFT model presented here alongside our dynamic approach towards modeling to optimize the design of digging and swimming robots across other design parameters of the origami-inspired fin (hinge distribution, stiffness, and geometry) \dl{and robot design such as size of the body and arm} in order to select designs and gaits that improve system speed, power, efficiency, etc based on a variety of needs \dl{and compensate the lift force in order to design a free-motion robot with supporting cart}. Another potential improvement is to implement rate-dependent RFT as in \cite{shashank_generalized_2020} to achieve more accurate results. 

\bibliographystyle{IEEEtran}
\balance
\bibliography{reference.bib}

\begin{thebibliography}{10}
\providecommand{\url}[1]{#1}
\csname url@samestyle\endcsname
\providecommand{\newblock}{\relax}
\providecommand{\bibinfo}[2]{#2}
\providecommand{\BIBentrySTDinterwordspacing}{\spaceskip=0pt\relax}
\providecommand{\BIBentryALTinterwordstretchfactor}{4}
\providecommand{\BIBentryALTinterwordspacing}{\spaceskip=\fontdimen2\font plus
\BIBentryALTinterwordstretchfactor\fontdimen3\font minus
  \fontdimen4\font\relax}
\providecommand{\BIBforeignlanguage}[2]{{%
\expandafter\ifx\csname l@#1\endcsname\relax
\typeout{** WARNING: IEEEtran.bst: No hyphenation pattern has been}%
\typeout{** loaded for the language `#1'. Using the pattern for}%
\typeout{** the default language instead.}%
\else
\language=\csname l@#1\endcsname
\fi
#2}}
\providecommand{\BIBdecl}{\relax}
\BIBdecl

\bibitem{hosoi_beneath_2015}
A.~Hosoi and D.~I. Goldman, ``Beneath {Our} {Feet}: {Strategies} for
  {Locomotion} in {Granular} {Media},'' \emph{Annual Review of Fluid
  Mechanics}, vol.~47, no.~1, pp. 431--453, 2015.

\bibitem{lauder_fish_2007}
G.~V. Lauder, E.~J. Anderson, J.~Tangorra, and P.~G.~A. Madden, ``Fish
  biorobotics: kinematics and hydrodynamics of self-propulsion,'' \emph{Journal
  of Experimental Biology}, vol. 210, no.~16, pp. 2767--2780, 2007.

\bibitem{dickinson_how_2000}
M.~H. Dickinson, ``How {Animals} {Move}: {An} {Integrative} {View},''
  \emph{Science}, vol. 288, no. 5463, pp. 100--106, 2000.

\bibitem{lauga_hydrodynamics_2009}
E.~Lauga and T.~R. Powers, ``The hydrodynamics of swimming microorganisms,''
  \emph{Reports on Progress in Physics}, vol.~72, no.~9, p. 096601, 2009.

\bibitem{purcell_life_1977}
E.~M. Purcell, ``Life at low {Reynolds} number,'' \emph{American Journal of
  Physics}, vol.~45, no.~1, pp. 3--11, 1977.

\bibitem{lucas_bending_2014}
K.~N. Lucas, N.~Johnson, W.~T. Beaulieu, E.~Cathcart, G.~Tirrell, S.~P. Colin,
  B.~J. Gemmell, J.~O. Dabiri, and J.~H. Costello, ``Bending rules for animal
  propulsion,'' \emph{Nature Communications}, vol.~5, no. May 2013, pp. 1--7,
  2014.

\bibitem{lauga_floppy_2007}
E.~Lauga, ``Floppy swimming: {Viscous} locomotion of actuated elastica,''
  \emph{Physical Review E}, vol.~75, no.~4, p. 041916, 2007.

\bibitem{peng_propulsion_2017}
Z.~Peng, Y.~Ding, K.~Pietrzyk, G.~J. Elfring, and O.~S. Pak, ``Propulsion via
  flexible flapping in granular media,'' \emph{Physical Review E}, vol.~96,
  no.~1, p. 012907, 2017.

\bibitem{peng_maximizing_2017}
Z.~Peng, G.~J. Elfring, and O.~S. Pak, ``Maximizing propulsive thrust of a
  driven filament at low {Reynolds} number via variable flexibility,''
  \emph{Soft Matter}, vol.~13, no.~12, pp. 2339--2347, 2017.

\bibitem{wang_dynamics_2018}
X.~Wang and S.~Alben, ``Dynamics and locomotion of flexible foils in a
  frictional environment,'' \emph{Proceedings of the Royal Society A:
  Mathematical, Physical and Engineering Sciences}, vol. 474, no. 2209, 2018.

\bibitem{mazouchova_flipper-driven_2013}
N.~Mazouchova, P.~B. Umbanhowar, and D.~I. Goldman, ``Flipper-driven
  terrestrial locomotion of a sea turtle-inspired robot,'' \emph{BIOINSPIRATION
  \& BIOMIMETICS}, vol.~8, no.~2, 2013.

\bibitem{marvi_sidewinding_2014}
H.~Marvi, C.~Gong, N.~Gravish, H.~Astley, D.~L. Hu, and D.~I. Goldman,
  ``Sidewinding with minimal slip: {Snake} and robot ascent ofsandy slopes,''
  \emph{Science}, vol. 346, no. 6206, 2014.

\bibitem{ortiz_soft_2019}
D.~Ortiz, N.~Gravish, and M.~T. Tolley, ``Soft {Robot} {Actuation} {Strategies}
  for {Locomotion} in {Granular} {Substrates},'' \emph{IEEE Robotics and
  Automation Letters}, vol.~4, no.~3, pp. 2630--2636, 2019.

\bibitem{https://doi.org/10.1002/aisy.201900183}
S.~Huang, Y.~Tang, H.~Bagheri, D.~Li, A.~Ardente, D.~Aukes, H.~Marvi, and J.~J.
  Tao, ``Effects of friction anisotropy on upward burrowing behavior of soft
  robots in granular materials,'' \emph{Advanced Intelligent Systems}, vol.~2,
  no.~6, p. 1900183, 2020.

\bibitem{Tao_2020}
J.~J. Tao, S.~Huang, and Y.~Tang, ``{SBOR}: a minimalistic soft
  self-burrowing-out robot inspired by razor clams,'' \emph{Bioinspiration {\&}
  Biomimetics}, vol.~15, no.~5, p. 055003, jul 2020.

\bibitem{doi:10.1061/(ASCE)GT.1943-5606.0002177}
J.~Tao, H.~Sichuan, and T.~Yong, ``Bioinspired self-burrowing-out robot in dry
  sand,'' \emph{Journal of Geotechnical and Geoenvironmental Engineering}, vol.
  145, no.~12, p. 02819002, 2019.

\bibitem{doi:10.1063/1.4802193}
K.~Shoele and Q.~Zhu, ``Performance of a wing with nonuniform flexibility in
  hovering flight,'' \emph{Physics of Fluids}, vol.~25, no.~4, p. 041901, 2013.

\bibitem{NIAN2020105944}
P.~Nian, B.~Song, J.~Xuan, W.~Zhou, and D.~Xue, ``Study on flexible flapping
  wings with three dimensional asymmetric passive deformation in a flapping
  cycle,'' \emph{Aerospace Science and Technology}, vol. 104, p. 105944, 2020.

\bibitem{tummala_design_2013}
Y.~Tummala, M.~Frecker, A.~Wissa, and J.~Hubbard, James~E., ``Design and
  optimization of a bend-and-sweep compliant mechanism,'' in \emph{IDETC}, 08
  2013.

\bibitem{mueller_incorporation_2009}
D.~Mueller, J.~W. Gerdes, and S.~K. Gupta, ``Incorporation of passive wing
  folding in flapping wing miniature air vehicles,'' in
  \emph{{IDETC}-{CIE}2009}.

\bibitem{russell_crabot_2011}
R.~A. Russell, ``{CRABOT}: {A} {Biomimetic} {Burrowing} {Robot} {Designed} for
  {Underground} {Chemical} {Source} {Location},'' \emph{Advanced Robotics},
  vol.~25, no.~1, pp. 119--134, 2011.

\bibitem{trueman_mechanism_1970}
E.~Trueman, ``The {Mechanism} of {Burrowing} of the {Mole} {Crab}, {Emerita},''
  \emph{Journal of Experimental Biology}, vol.~53, no.~3, pp. 701--710, 1970.

\bibitem{onal_towards_2011}
C.~D. Onal, R.~J. Wood, and D.~Rus, ``Towards printable robotics:
  {Origami}-inspired planar fabrication of three-dimensional
  mechanisms.''\hskip 1em plus 0.5em minus 0.4em\relax IEEE, 2011, pp.
  4608--4613.

\bibitem{6266749}
D.~Onal, Cagdas, J.~Wood, Robert, and D.~Rus, ``An origami-inspired approach to
  worm robots,'' \emph{IEEE/ASME Transactions on Mechatronics}, vol.~18, no.~2,
  pp. 430--438, 2013.

\bibitem{wood_first_2008}
R.~Wood, ``The {First} {Takeoff} of a {Biologically} {Inspired} {At}-{Scale}
  {Robotic} {Insect},'' \emph{IEEE Transactions on Robotics}, vol.~24, no.~2,
  pp. 341--347, 2008.

\bibitem{tolley_self-folding_2014}
M.~T. Tolley, S.~M. Felton, S.~Miyashita, D.~Aukes, D.~Rus, and R.~J. Wood,
  ``Self-folding origami: shape memory composites activated by uniform
  heating,'' \emph{Smart Materials and Structures}, vol.~23, no.~9, p. 094006,
  2014.

\bibitem{doi:10.1089/soro.2019.0156}
M.~Jiang, Z.~Zhou, and N.~Gravish, ``Flexoskeleton printing enables versatile
  fabrication of hybrid soft and rigid robots,'' \emph{Soft Robotics}, 2020.

\bibitem{maladen_undulatory_2009}
R.~D. Maladen, Y.~Ding, C.~Li, and D.~I. Goldman, ``Undulatory {Swimming} in
  {Sand}: {Subsurface} {Locomotion} of the {Sandfish} {Lizard},''
  \emph{Science}, vol. 325, no. 5938, pp. 314--318, 2009.

\bibitem{gray_propulsion_1955}
B.~Y.~J. Gray and G.~J. Hancockf, ``The {Propulsion} of {Sea}-{Urchin}
  {Spermatozoa},'' \emph{Journal of Experimental Biology}, vol.~32, no.~4, pp.
  802--814, 1955.

\bibitem{maladen_mechanical_2011}
R.~D. Maladen, Y.~Ding, P.~B. Umbanhowar, A.~Kamor, and D.~I. Goldman,
  ``Mechanical models of sandfish locomotion reveal principles of high
  performance subsurface sand-swimming,'' \emph{Journal of the Royal Society
  Interface}, vol.~8, no.~62, pp. 1332--1345, 2011.

\bibitem{goldman_colloquium_2014}
D.~I. Goldman, ``Colloquium: {Biophysical} principles of undulatory
  self-propulsion in granular media,'' \emph{Reviews of Modern Physics},
  vol.~86, no.~3, pp. 943--958, 2014.

\bibitem{li_terradynamics_2013}
C.~Li, T.~Zhang, and D.~I. Goldman, ``A {Terradynamics} of {Legged}
  {Locomotion} on {Granular} {Media},'' \emph{Science}, vol. 339, no. 6126, pp.
  1408--1412, 2013.

\bibitem{zhang_effectiveness_2014}
T.~Zhang and D.~I. Goldman, ``The effectiveness of resistive force theory in
  granular locomotion,'' \emph{Physics of Fluids}, vol.~26, no.~10, pp. 0--17,
  2014.

\bibitem{sharifzadeh_integrated_2018}
M.~Sharifzadeh, R.~Khodambashi, and D.~M. Aukes, ``An integrated design and
  simulation environment for rapid prototyping of laminate robotic
  mechanisms,'' in \emph{{IDETC}-2018}.

\bibitem{9369911}
M.~Sharifzadeh, Y.~Jiang, and D.~M. Aukes, ``Reconfigurable curved beams for
  selectable swimming gaits in an underwater robot,'' \emph{IEEE Robotics and
  Automation Letters}, vol.~6, no.~2, pp. 3437--3444, 2021.

\bibitem{rapaport_art_2004}
D.~C. Rapaport, \emph{The {Art} of {Molecular} {Dynamics} {Simulation}}.\hskip
  1em plus 0.5em minus 0.4em\relax Cambridge University Press, 2004.

\bibitem{michael_demfem_2015}
M.~Michael, F.~Vogel, and B.~Peters, ``{DEM}–{FEM} coupling simulations of
  the interactions between a tire tread and granular terrain,'' \emph{Computer
  Methods in Applied Mechanics and Engineering}, vol. 289, pp. 227--248, 2015.

\bibitem{tu_sequential_2017}
F.~Tu, D.~Delbergue, H.~Miao, T.~Klotz, M.~Brochu, P.~Bocher, and M.~Levesque,
  ``A sequential {DEM}-{FEM} coupling method for shot peening simulation,''
  \emph{Surface and Coatings Technology}, vol. 319, pp. 200--212, 2017.

\bibitem{han_combined_2000}
K.~Han, D.~Peric, D.~Owen, and J.~Yu, ``A combined finite/discrete element
  simulation of shot peening processes – {Part} {II}: {3D} interaction
  laws,'' \emph{Engineering Computations}, vol.~17, no.~6, pp. 680--702, 2000.

\bibitem{wieghardt_experiments_1975}
K.~Wieghardt, ``Experiments in {Granular} {Flow}.'' \emph{Annu Rev Fluid Mech},
  vol.~v, pp. 89--114, 1975.

\bibitem{PhysRevE.64.061303}
I.~Albert, J.~G. Sample, A.~J. Morss, S.~Rajagopalan, A.-L. Barab\'asi, and
  P.~Schiffer, ``Granular drag on a discrete object: Shape effects on
  jamming,'' \emph{Phys. Rev. E}, vol.~64, p. 061303, Nov 2001.

\bibitem{albert_slow_1999}
R.~Albert, M.~A. Pfeifer, A.~L. Barabási, and P.~Schiffer, ``Slow drag in a
  granular medium,'' \emph{Physical Review Letters}, vol.~82, no.~1, pp.
  205--208, 1999.

\bibitem{hansen2019pycma}
N.~Hansen, Y.~Akimoto, and P.~Baudis, ``{CMA-ES/pycma} on {G}ithub,'' Zenodo,
  DOI:10.5281/zenodo.2559634, Feb. 2019.

\bibitem{ding_drag_2011}
Y.~Ding, N.~Gravish, and D.~I. Goldman, ``Drag induced lift in granular
  media,'' \emph{Physical Review Letters}, vol. 106, no.~2, pp. 1--4, 2011.

\bibitem{maladen_granular_2011}
R.~D. Maladen, P.~B. Umbanhowar, Y.~Ding, A.~Masse, and D.~I. Goldman,
  ``Granular lift forces predict vertical motion of a sand-swimming robot,''
  2011, pp. 1398--1403.

\bibitem{shashank_generalized_2020}
A.~Shashank, K.~Andras, G.~D. I, and K.~Ken, ``A generalized resistive force
  theory for rate-dependent intrusion phenomena in granular media,'' 2020.

\end{thebibliography}

\end{document}